\documentclass[5p,twocolumn]{elsarticle}

\usepackage{lineno,hyperref}
\modulolinenumbers[5]

\usepackage{times}
\usepackage{booktabs}
\usepackage{multirow}
\usepackage{threeparttable} 

\journal{Optical Switching and Networking}

\begin{document}
	
	\begin{frontmatter}
		
		\title{Artificial Intelligence (AI) Methods in Optical Networks: A Comprehensive Survey}
		
		\author[UVa]{Javier Mata}
		\ead{javier.mata@ribera.tel.uva.es}
		\author[UVa]{Ignacio de Miguel}
		\ead{ignacio.miguel@tel.uva.es}
		\author[UVa]{Ram\'{o}n J. Dur\'{a}n}
		\author[UVa]{Noem\'{i} Merayo}
		\author[TUBS]{Sandeep Kumar Singh}
		\author[TUBS]{Admela Jukan}
		\author[ADVA]{Mohit Chamania}
		\address[UVa]{Universidad de Valladolid, ETSI de Telecomunicaci\'{o}n, Campus Miguel Delibes, 47011 Valladolid, Spain}
		\address[TUBS]{Technische Universit\"{a}t Carolo-Wilhelmina zu Braunschweig, Germany}
		\address[ADVA]{ADVA Optical Networking, Berlin, Germany}

		\begin{abstract}
			Artificial intelligence (AI) is an extensive scientific discipline which enables computer systems to solve problems by emulating complex biological processes such as learning, reasoning and self-correction. This paper presents a comprehensive review of the application of AI techniques for improving performance of optical communication systems and networks. The use of AI-based techniques is first studied in applications related to optical transmission, ranging from the characterization and operation of network components to performance monitoring, mitigation of nonlinearities, and quality of transmission estimation. Then, applications related to optical network control and management are also reviewed, including topics like optical network planning and operation in both transport and access networks. Finally, the paper also presents a summary of opportunities and challenges in optical networking where AI is expected to play a key role in the near future.
		\end{abstract}
		
		
	\end{frontmatter}
	
	
	\nocite{russell2009artificial, russell2009artificial, mukherjee2006optical, simmons2014optical, rao2012teachinglearning, du2016search, martinelli2014genetic, Morgan2015Ant, Monoyios2011Multiobjective,Wang2014Distributed, awwad2009bayesian, zhu2017leveraging, pavel2012game, de2013cognitive, borkowski2015cognitive, zervas2010cognitive, wei2012cognitive, yoo2014multi-domain, chan2017cognitive, gosselin2017application, chitra2014hidden, szafraniec2013performance, yannuzzi2016interdomain, jayaraj2008loss, tachibana2010mdp-based, sue2011dynamic, reyes2017adaptive, russell2009artificial, ghahramani2015probabilistic, Murphy2012Machine, bishop2006pattern, taylor2009phase, karinou2017solutions, rottenberg2017ml, russell2009artificial, wu2009applications, jimenez2012cognitive, jimenez2013cognitive, caballero2012experimental, barletta2017qot, oda2017learning, Rastegarfar2016TCP, russell2009artificial, gonzalez2010cognitive, tan2014simultaneous, torres2016adaptive, kaelbling1996reinforcement, watkins1992q, kiran2007reinforcement, zibar2015application, hraghi2017demonstration, brunton2014self-tuning, huang2016machine, huang2017dynamic, barboza2013self, szafraniec2013performance, wu2009applications, tan2014simultaneous, tanimura2016osnr, thrane2017machine, tan2014simultaneous, thrane2017machine, gonzalez2010cognitive, taylor2009phase, karinou2017solutions, lau2007signal, rottenberg2017ml, zibar2016machine, wang2015nonlinear, wang2016knn, torres2016adaptive, giacoumidis2017nonlinear, jimenez2012cognitive, jimenez2013cognitive, caballero2012experimental, jimenez2013cognitive, barletta2017qot, oda2017learning, Mata2017SVM, duthel2009laser,kikuchi2012characterization, zibar2015application, hraghi2017demonstration, brunton2014self-tuning, huang2016machine, huang2017dynamic, barboza2013self, dong2016optical, wu2009applications, szafraniec2013performance, wu2009applications, li2010signed,khan2010osnr, tan2014simultaneous, tanimura2016osnr, thrane2017machine, lau2007signal, napoli2014reduced, irukulapati2016stochastic, gonzalez2010cognitive, zibar2016machine, wang2015nonlinear, wang2016knn, torres2016adaptive, giacoumidis2017nonlinear, azodolmolky2011experimental, jimenez2012cognitive, aamodt1994case, jimenez2012cognitive, jimenez2013cognitive, caballero2012experimental, Mata2017SVM, jimenez2013cognitive, barletta2017qot, oda2017learning, morais2011genetic, Morgan2015Ant, martinelli2014genetic, Cerutti2014Trading, Morgan2015Ant, Velasco2013Saving, demiguel2009genetic, Durand2013Energy, Paula2016WDM, Petridou2008Clustering, reyes2017adaptive, rubio2012comparative, martinelli2014genetic, Monoyios2011Multiobjective, gong2012two-population, lechowicz2016genetic, Wright2015Minimum, przewozniczek2015towards, Wang2014Distributed, kyriakopoulos2016energy, Yuanyuan2016Optimized, Chen2014Cognitive, christodoulopoulos2011elastic, perello2016flex, walkowiak2014routing, goscien2015tabu, Jia2016Efficient, kiran2007reinforcement, zhu2017leveraging, Araujo2015Methodology, yannuzzi2016interdomain, tachibana2010mdp-based, sue2011dynamic, Corut2010Ensuring, fernandez2012survivable, fernandez2013techno, Gao2015Virtual, fernandez2015virtual, morales2017virtual, ruiz2016service, Zhang2016Failure, gosselin2017application, Tembo2016Tutorial, Montero2017Dynamic, oliveira2015toward, Yoo2014Multi, Yan2017Field, praveen2006first, kiran2007reinforcement, venkatesh2008complete, Haeri2013Reinforcement, jayaraj2008loss, Elbiaze2011Cognitive, Haeri2015Intelligent, Leung2017Extreme, coulibaly2015qos-aware, chitra2014hidden, awwad2009bayesian, oda2017learning, Villalba2009Design, Kokangul2011Optimization, Liu2012Nested, Huang2008Study, Moradpoor2013Mathematical, Jimenez2015Auto, Hwang2012Genetic, Merayo2017PID, Dias2014Bayesian, Bhatt2017ONU, Bhatt2015Hybrid, de2013cognitive, borkowski2015cognitive, zervas2010cognitive, wei2012cognitive, yoo2014multi-domain, chan2017cognitive, Rastegarfar2016TCP, Glick2017Scheduling, Wang2017Adaptive, mukherjee2006optical, simmons2014optical, du2016search, morais2011genetic, demiguel2009genetic, martinelli2014genetic, Morgan2015Ant, Cerutti2014Trading, Velasco2013Saving, Durand2013Energy, Paula2016WDM, Petridou2008Clustering, Monoyios2011Multiobjective, rubio2012comparative, Wang2014Distributed, Chen2014Cognitive, kyriakopoulos2016energy, Yuanyuan2016Optimized, Araujo2015Methodology, Klinkowski2011Routing, christodoulopoulos2011elastic, perello2016flex, gong2012two-population, lechowicz2016genetic, walkowiak2014routing, przewozniczek2015towards, goscien2015tabu, aibin2014simulated, Jia2016Efficient, Wright2015Minimum, Corut2010Ensuring, fernandez2012survivable, fernandez2013techno, fernandez2015virtual, ruiz2016service, morales2017virtual, Gao2015Virtual, Xia2015Survey, Montero2017Dynamic, Zhang2016Failure, oliveira2015toward, Yan2017Field, Yoo2014Multi, praveen2006first, praveen2006first, kiran2007reinforcement, venkatesh2008complete, jayaraj2008loss, Leung2017Extreme, Elbiaze2011Cognitive, Haeri2013Reinforcement, Haeri2015Intelligent, coulibaly2015qos-aware, Villalba2009Design, Kokangul2011Optimization, Liu2012Nested, Bhatt2017ONU, Bhatt2015Hybrid, gosselin2017application, Tembo2016Tutorial, Sarigiannidis2017DIANA, Huang2008Study, Hwang2012Genetic, Moradpoor2013Mathematical, Jimenez2015Auto, Merayo2017PID, Dias2014Bayesian, Rastegarfar2016TCP, Glick2017Scheduling, Wang2017Adaptive} 
		

	\section{Introduction}
	
	Artificial intelligence (AI) entities and systems have the ability to perform operations analogous to learning and decision making by imitating biological processes, with special emphasis on human cognitive processes. AI applications such as virtual personal-assistants, smart vehicles, purchase prediction, speech recognition or smart home devices, are almost ubiquitous, and similar AI-based techniques are already changing our daily lives in ways that improve human productivity, safety or health, affecting even the way we entertain or communicate. 
	
	For the most part, AI does not deliver completely autonomous systems, but instead adds knowledge and reasoning to existing applications, databases, and environments, to make them friendlier, smarter, and more sensitive to changes in their environments. Each small breakthrough on AI research enables us to expand our skills to solve new classes and scales of problems, thereby driving research and innovation in almost every scientific discipline.
	
	As an example, the improvement of the performance of telecommunication networks by the application of AI-based techniques has become an area under extensive research over the past decades, affecting areas of transmission, switching and network management. Optical communication networks and systems have not stayed on the sidelines, but have started to adopt this discipline towards AI-based optical networking, from photonic devices to control and management.
	
	The aim of this paper is to review some of the currently considered approaches to increase the performance of optical networks by the use of AI mechanisms, providing a survey of the current research within this area, as well as an overview of opportunities and challenges arising in this context.
	
	The remainder of this paper is organized as follows. Section \ref{sec:AI-techniques} provides an introduction to the field of AI. Since that is a very broad area, we review those AI subfields --and their associated techniques-- which have had or are expected to have, in our opinion, a significant role in optical networking. Then, Sections \ref{sec:role-AI-physical} and \ref{sec:role-AI-networks} analyze the role of AI techniques in optical communication systems and networks. We first survey the use of AI in optical transmission (Section \ref{sec:role-AI-physical}), and then we focus on networking issues (Section \ref{sec:role-AI-networks}). Finally, in Section \ref{sec:opportunities} we describe further opportunities and challenges, and we conclude in Section \ref{sec:conclusion}.
	
	\section{An Overview of AI and Related Techniques} \label{sec:AI-techniques}
	
	AI focuses on the study of intelligent or rational agents, i.e., entities which perceive and act in an environment with the aim of achieving their goals or maximizing a performance parameter. Moreover, they can further improve their performance through learning \cite{russell2009artificial}.
	
\begin{figure*}
		\begin{center} 
			\includegraphics[width=18cm]{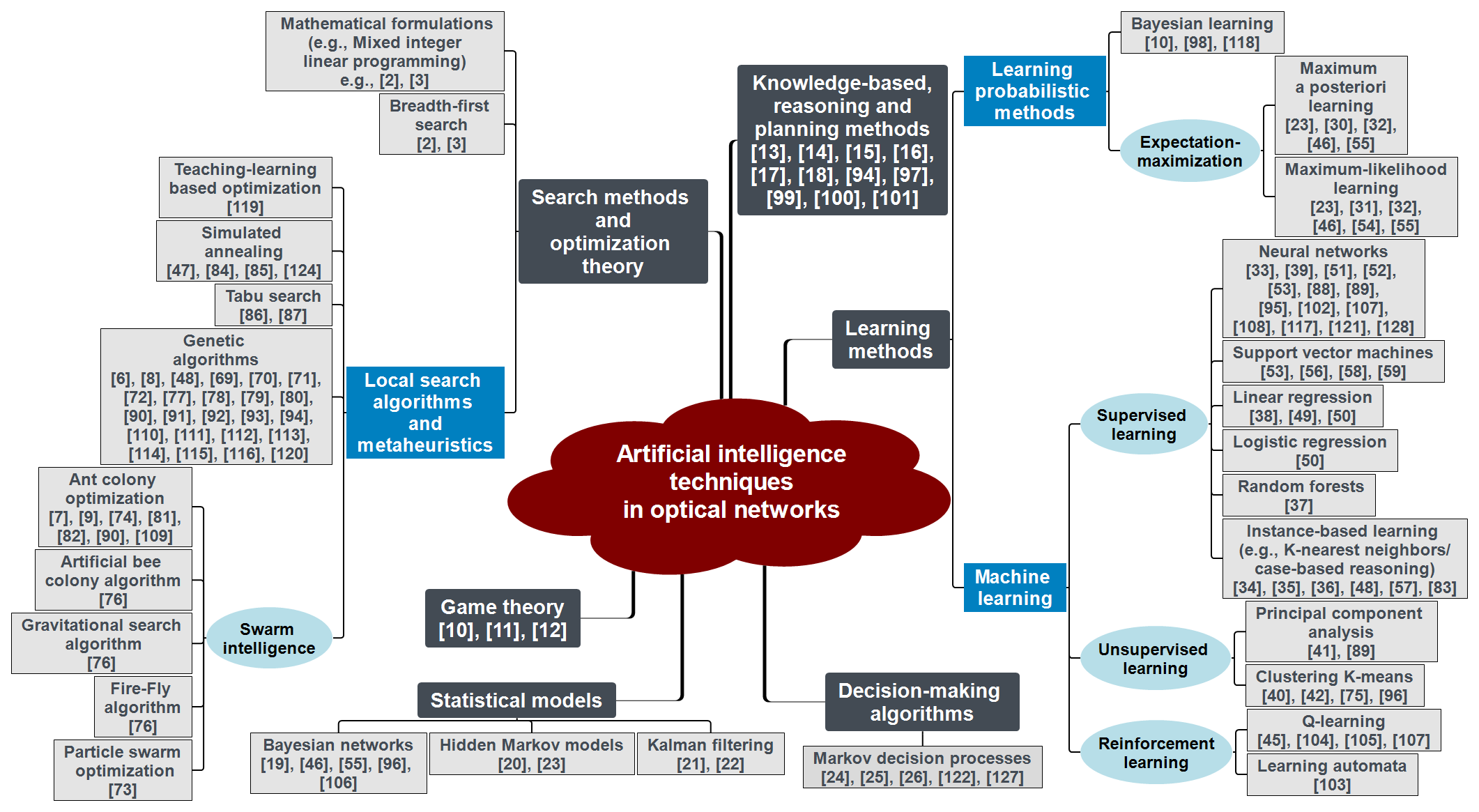}
			\caption{AI subfields and techniques applied to optical networks}
			\label{figure:AI-opticalnetworks}
		\end{center}
	\end{figure*}

	In this section, we briefly go through some of the subfields of AI that have been successfully employed in optical networking, stating the motivation for their introduction, and providing some examples of their use in the optical networking literature. Figure \ref{figure:AI-opticalnetworks} shows a diagram with AI subfields and techniques, and classifies the references reviewed in this survey within those categories.

	The simplest type of networking scenarios that we can think of are deterministic, observable, static and completely known. For these scenarios, \textit{search algorithms and optimization theory} are key elements of the AI area \cite{russell2009artificial}, and thus, they have been extensively used in optical network design and control for a long time. Examples include the use of breadth-first-search algorithms for routing, and linear and mixed-integer linear programming formulations for network planning (e.g., \cite{mukherjee2006optical, simmons2014optical}). However, when some of the conditions previously mentioned are relaxed, or when the network size prohibits the use of the former techniques, those methods have been complemented or replaced by \textit{local search algorithms and metaheuristics} like simulated annealing, genetic algorithms, swarm optimization, and teaching-learning based optimization \cite{rao2012teachinglearning, du2016search}. For instance, optical network planning \cite{martinelli2014genetic, Morgan2015Ant} and lightpath establishment \cite{Monoyios2011Multiobjective,Wang2014Distributed} have benefited from those techniques.
	
	In many cases, the optical network has a single point of intelligence (like a centralized control node), i.e., a single agent. However, in other cases, different intelligent agents are involved, so that the actions taken by an agent have an impact on the others. For those scenarios, \textit{game theory} may come into play, and proposals in the area of optical networking can be found, e.g., in  \cite{awwad2009bayesian} (in the context of hybrid radio-frequency/free space optics networks), \cite{zhu2017leveraging} (in the context of elastic optical networks, EONs), or \cite{pavel2012game} (a book completely devoted to the topic).
	
	A stride forward towards making agents more intelligent comes by incorporating the use of \textit{knowledge, reasoning and planning}. In this case, intelligent agents keep a knowledge base (KB) where relevant knowledge about the environment and about the impact of their actions is stored. That KB is used by the agents when devising plans of action on how to succeed on decision-making, and can be updated in order to adapt to changing conditions. Along this line, holistic frameworks, like cognitive optical networks, which perceive, act, learn, adapt and optimize their performance, have been proposed by different researchers \cite{de2013cognitive, borkowski2015cognitive, zervas2010cognitive, wei2012cognitive, yoo2014multi-domain, chan2017cognitive}. 
	
	Three noteworthy topics when it comes to incorporating intelligence to optical systems or networks are \textit{how to handle uncertainty}, \textit{how to tackle decision-making}, and \textit{how to learn}. 
	
	For sure, in an optical network there are non-deterministic events taking place, and lack of full information about the environment is not a rare issue. Therefore, intelligent agents must be able to operate under uncertainty in a robust way. The laws of probability and, in particular, \textit{Bayesian networks} are useful tools to build those robust models (e.g. \cite{gosselin2017application}). Moreover, optical systems and networks are subject to constant changes. Hence, intelligent agents must include inference algorithms for temporal models to perform tasks like filtering, prediction or smoothing, relying on techniques like \textit{hidden Markov models (HMM)} \cite{chitra2014hidden} and \textit{Kalman filters} \cite{szafraniec2013performance, yannuzzi2016interdomain}. 
	
	A second key element is the use of decision-making algorithms. The underlying principle for these algorithms is the maximization of the expected utility, in which a utility function is defined in order to assign a single number to express the desirability of a state and an agent makes decisions with the aim of maximizing such a function (e.g. \cite{jayaraj2008loss}). Realistic networking environments, however, must deal with uncertainty and the utility of an agent usually depends on a sequence of decisions rather than on a single isolated one. Decision making in optical network agents can therefore be modelled as sequential decision problems in uncertain environments. These problems can be solved by \textit{Markov decision processes (MDPs)} if the agent's actions depend only on the current state of the agent, and not on its history. MDPs are defined by a transition model, which specifies the probabilistic outcomes of actions, and by a reward function, which specifies the reward in each state. The solution of an MDP is a policy that associates a decision with every state that the agent might reach. An optimal policy maximizes the utility of the state sequences encountered when it is executed. The use of MDPs in optical networking has been shown in \cite{tachibana2010mdp-based, sue2011dynamic, reyes2017adaptive, baldine2001traffic}.
	
	The third issue of paramount importance is learning. Learning enables an agent to improve its performance on future tasks due to acquired experience. The inclusion of learning is important for several reasons. A learning-capable agent can adapt to changes in the environment and it is even able to adapt to unforeseen scenarios that could not be anticipated when the agent was designed. Moreover, in many cases, learning from existing data may be the only way to generate a working model, or in other words, as stated in \cite{russell2009artificial}, sometimes human programmers (or engineers) have no idea on how to program a solution themselves. Statistical learning and machine learning provide the theory and tools to learn from existing data, which can be gathered in optical communications systems and networks thanks to monitoring techniques.
	
	Although agents can handle uncertainty by using the methods of probability and decision theory, they must learn their probabilistic theories from experience. Thus, \textit{bayesian learning methods} \cite{ghahramani2015probabilistic} formulate learning as a form of probabilistic inference, using the observations to update a prior distribution over hypotheses; \textit{maximum a posteriori (MAP) learning} \cite{Murphy2012Machine} selects a single most likely hypothesis given the data, and \textit{maximum-likelihood learning} \cite{bishop2006pattern} simply selects the hypothesis that maximizes the likelihood of the data. These techniques have been used in optical receivers, e.g., in \cite{taylor2009phase, karinou2017solutions, rottenberg2017ml}.
	
	Apart from the above mentioned techniques \textit{machine learning} has also been widely used. There are three main categories in \textit{machine learning}. In \textit{supervised learning} \cite{russell2009artificial} an agent observes some example input-output pairs and learns a function that maps from input to output. Techniques include linear regression, logistic regression, decision trees, artificial neural networks, nearest neighbor models and support vector machines (SVM) to name just a few. Moreover, different models can be combined in ensemble learning, with the aim of improving results. Supervised learning has been used, for instance, for optical performance monitoring \cite{wu2009applications}, to estimate the quality of transmission (QoT) in optical networks \cite{jimenez2012cognitive, jimenez2013cognitive, caballero2012experimental, barletta2017qot, oda2017learning} and for resource allocation in data centers \cite{Rastegarfar2016TCP}. In \textit{unsupervised learning} \cite{russell2009artificial}, an agent learns patterns from the input even though no explicit output is supplied. For instance, clustering and principal component analysis methods, which belong to this type of learning, have been used for optical performance monitoring, modulation format recognition and impairment mitigation \cite{gonzalez2010cognitive, tan2014simultaneous, torres2016adaptive}. Finally, in \textit{reinforcement learning} \cite{kaelbling1996reinforcement} an agent learns an optimal (or nearly optimal) policy from a series of reinforcements (rewards) or punishments received from its interaction with the environment. Some techniques include adaptive dynamic programming and temporal-difference (TD) methods. Q-learning, a well-known technique of the latter type, aims to find an optimal quality value (Q-Value) of action-selection policy for any given (finite) Markov decision process \cite{watkins1992q}. For instance, Q-learning has been used for path and wavelength selection in the context of optical burst-switched (OBS) networks \cite{kiran2007reinforcement}.
	
	\section{Applications of AI in Optical Transmission} \label{sec:role-AI-physical} 
	
	\begin{table*}[ht!]
		\centering
		\scriptsize
		\caption{Applications in optical transmission taking advantage of AI techniques}
		\label{table:AI-physical}
		\begin{tabular}{@{}lll@{}}
			\toprule
			\midrule
			\multicolumn{1}{l}{\textbf{Applications} }                                                     & \multicolumn{1}{l}{\textbf{AI techniques} }& \multicolumn{1}{c}{\textbf{Literature}                                                                                                                                                                                                                         } \\ \midrule 
			\multirow{5}{*}{Transmitters} &\begin{tabular}[c]{@{}l@{}} Bayesian filtering and\\ expectation-maximization \end{tabular}& \begin{tabular}[c]{@{}l@{}} \cite{zibar2015application}: characterizes laser amplitude and phase noise. \end{tabular} \\ & Simulated annealing &\begin{tabular}[c]{@{}l@{}} \cite{hraghi2017demonstration}: determines the optimal settings for optical comb sources for ultradense WDM passive optical networks.\end{tabular}           \\ & \begin{tabular}[c]{@{}l@{}} Machine learning (pattern\\ learning methods) and genetic\\ algorithms\end{tabular}                     &\begin{tabular}[c]{@{}l@{}} \cite{brunton2014self-tuning}: self-tuning mechanism for mode-locked fiber lasers. \end{tabular}                               \\ \midrule
			Optical amplification control                                                     & \begin{tabular}[c]{@{}l@{}} Kernelized linear regression  \\  Linear/logistic regression \\ Multilayer perceptron neural network  \end{tabular}                        & \begin{tabular}[c]{@{}l@{}} \cite{huang2016machine}: defines regression model to study power excursions in multi-span EDFA networks. \\ \cite{huang2017dynamic}: uses a ridge regression model to cope with the discrepancy among post-EDFA channel powers. \\ \cite{barboza2013self}: autonomous adjustment of the operating point of amplifiers in an EDFA cascade.
			\end{tabular} \\ \midrule
			Linear impairments identification & \begin{tabular}[c]{@{}l@{}}Kalman filter\\ Neural networks\\Principal component analysis \end{tabular} & \begin{tabular}[c]{@{}l@{}}\cite{szafraniec2013performance}: carrier phase tracking, polarization tracking, and estimation of the first-order PMD. \\ \cite{wu2009applications}: identifies CD, PMD and OSNR provided that bit-rate and modulation format is known.\\ \cite{tan2014simultaneous}: monitors CD, PMD and OSNR. \end{tabular} \\ \midrule 
			\multirow{1}{*}{OSNR monitoring}                                            & \begin{tabular}[l]{@{}l@{}}Deep neural networks (DNN)\\ Neural networks   \end{tabular}                    & \begin{tabular}[c]{@{}l@{}} \cite{tanimura2016osnr}: uses DNN,  trained with asynchronously sampled  raw data,  for OSNR monitoring.\\\cite{thrane2017machine}: uses neural networks based nonlinear regression for OSNR estimation. \end{tabular}                               \\ \midrule
			Modulation format recognition                                                      & \begin{tabular}[c]{@{}l@{}}Principal component analysis\\ Support vector machines (SVM)\\ Clustering k- means \end{tabular}                         & \begin{tabular}[c]{@{}l@{}}\cite{tan2014simultaneous}: identifies modulation formats/bit rates from a known set.\\ \cite{thrane2017machine}: classifies modulation formats using the variance of  eye opening width.    \\  \cite{gonzalez2010cognitive}: identifies modulation formats based on the number of levels and clusters in constellation diagram. \end{tabular} \\ \midrule
			\multirow{13}{*}{Receivers, nonlinearity mitigation}   &
			\begin{tabular}[c]{@{}l@{}}Maximum a posteriori\end{tabular}                  &\begin{tabular}[c]{@{}l@{}}\cite{taylor2009phase}: looks for phase estimates feasible to calculate in real-time.\end{tabular}\\&
			\begin{tabular}[c]{@{}l@{}}Maximum-likelihood\end{tabular}      &\begin{tabular}[c]{@{}l@{}}\cite{karinou2017solutions}: proposes various equalization schemes for high capacity WDM interconnects.\end{tabular}\\&&

			 \begin{tabular}[c]{@{}l@{}}\cite{lau2007signal}: Maximum-likelihood detection for phase-modulated systems with linear and nonlinear phase noise.\end{tabular}  \\            &
			 			\begin{tabular}[c]{@{}l@{}}Maximum-likelihood and\\ maximum a posteriori\end{tabular}    &\begin{tabular}[c]{@{}l@{}}\cite{rottenberg2017ml}: proposes various estimators to recover the phase in Offset-QAM-based filterbank multicarrier systems.\end{tabular}\\
			                   & \begin{tabular}[c]{@{}l@{}}Bayesian filtering and\\ expectation-maximization \end{tabular}  & \begin{tabular}[c]{@{}l@{}} \cite{zibar2016machine}: proposes state-space models for cross-polarization mitigation, carrier synchronization, symbol \\ \,\,\,\,\,\,\,\,\,\,\,\,\,\,  detection.\end{tabular}  \\                                                                          
			& \begin{tabular}[c]{@{}l@{}}Nonlinear support vector\\ machines\end{tabular}  & \begin{tabular}[c]{@{}l@{}}\cite{wang2015nonlinear}: SVM is applied to create decision boundaries to avoid  errors induced by  nonlinear impairment.\end{tabular}                                                                                                        \\
			& K-nearest neighbors &\begin{tabular}[c]{@{}l@{}} \cite{wang2016knn}: proposes an algorithm that learns the link properties and generates the nonlinear decision boundaries for \\ \,\,\,\,\,\,\,\,\,\,\,\,\,\, maximizing transmission distance and improving nolinear tolerance.\end{tabular}  \\ 
			& Clustering k-means & \cite{torres2016adaptive}: proposes a technique to mitigate the effect of time-varying impairments, e.g., phase noise. \\ 
			&\begin{tabular}[c]{@{}l@{}} Nonlinear support vector\\ machines and Newton method\end{tabular}  & \cite{giacoumidis2017nonlinear}: uses Newton-method (N-SVM) to reduce inter-subcarrier nonlinear crosstalk effects. \\ \midrule
			
			\multirow{6}{*}{QoT estimation}                                          & Case-Based Reasoning (CBR) &\begin{tabular}[c]{@{}l@{}}  \cite{jimenez2012cognitive}: presents a QoT estimator to decide whether a lightpath fulfils QoT requirements or not. \end{tabular} \\ &   CBR + learning/forgetting              & \begin{tabular}[c]{@{}l@{}} \cite{jimenez2013cognitive}: optimizes previous CBR approach for QoT estimation with learning and forgetting techniques.      \end{tabular}      \\    
			&   CBR + learning/forgetting              & \begin{tabular}[c]{@{}l@{}} \cite{caballero2012experimental}: experimental demonstration of the QoT estimator \cite{jimenez2013cognitive} in a WDM 80 Gb/s PDM-QPSK testbed.      \end{tabular}      \\  
			& Random forests classifier  & \begin{tabular}[c]{@{}l@{}}\cite{barletta2017qot}: predicts the probability that the BER of a candidate lightpath will not exceed a given threshold.   \end{tabular} \\
			& Linear regression &\begin{tabular}[c]{@{}l@{}}\cite{oda2017learning}: uses BER information monitoring and a learning process (based on linear regression)\\ \,\,\,\,\,\,\,\,\,\,\,\,\,\,  in order to estimate the BER of each new service request.\end{tabular}\\
				& Support vector machines &\begin{tabular}[c]{@{}l@{}}\cite{Mata2017SVM}: proposes a fast and accurate lightpath QoT estimator based on SVM to decide whether a lightpath fulfils \\ \,\,\,\,\,\,\,\,\,\,\,\,\,\, QoT requrements or not.\end{tabular} 
			\\ \midrule
			\bottomrule
		\end{tabular}
	\end{table*}
	
	In this section, we describe applications of AI techniques in the physical layer of optical networks, i.e., in optical transmission-related issues. AI techniques can help improve the configuration and operation of network devices, optical performance monitoring, modulation format recognition, fiber nonlinearities mitigation and quality of transmission (QoT) estimation. These applications are summarized in Table \ref{table:AI-physical}.  
	
	\subsection{Characterization and Operation of Transmitters}
	
	AI techniques facilitate statistical modeling of individual optical components by including the underlying physics. In all these cases where a deterministic approach results in an impractical computational load, learning mechanisms are becoming a promising and accurate performance improvement tool.
	
	With the advent of advanced modulation formats aiming to increase the spectral efficiency, ranging from 16 quadrature amplitude modulation (16 QAM) to 64 QAM and beyond, the need for robust carrier frequency and phase synchronization becomes crucial. At this point, a precise characterization of amplitude and phase noise of lasers is essential. Conventional time-domain approaches perform coherent detection in combination with digital signal processing (DSP) to cope with this issue \cite{duthel2009laser,kikuchi2012characterization}, but as higher order modulation formats are implemented, the accuracy of the phase noise estimation is compromised in the presence of moderate measurement noise. Zibar \textit{et al}.\ \cite{zibar2015application} present a framework of Bayesian filtering in combination with expectation maximization (EM) to accurately characterize laser amplitude and phase noise that outperforms conventional approaches. Results demonstrate an accurate estimation of the phase noise even in the presence of large measurement noise.
	
	Additional examples of the use of AI techniques in the optimization of transmitters and lasers include the work by Hragui \textit{et al}.\ \cite{hraghi2017demonstration}, who use simulated annealing to determine the optimal settings in terms of flatness for optical comb sources for ultradense WDM passive optical networks, and the work by Brunton \textit{et al}.\ \cite{brunton2014self-tuning}, who jointly use machine learning, genetic algorithms and adaptive control techniques to provide a self-tuning mechanism for mode-locked fiber lasers.

	\subsection{Operation of Erbium-Doped Fiber Amplifiers (EDFAs)}
	
	EDFAs are another optical network component on which AI techniques have been extensively applied. EDFAs are one of the key elements of optical transport networks, capable of extending the reach of the transmitted optical signal by performing amplification of WDM channels in the optical domain. Machine learning techniques offer efficient solutions to a wide range of challenges inherent to the operation of EDFAs within optical fiber transmission. 
	
	Specifically, Huang \textit{et al}.\ \cite{huang2016machine} define a regression problem with supervised machine learning (using a radial basis function) to statistically model the channel dependence of power excursions in multi-span EDFA networks, learning from historical data. It provides the system with accurate recommendations on channel add/drop strategies to minimize the power disparity among channels. With the arrival of flex-grid networks, in which dynamic defragmentation is often applied to reoptimize spectrum assignment to active connections in order to improve the spectral efficiency, the previous study is extended in \cite{huang2017dynamic} to cope with the power excursion problem in dynamically changing spectral configurations. A ridge regression model is used to determine the magnitude of the impact of a given sub-channel, and a logistic regression is applied to specify whether the contribution will result in an increase or decrease in the discrepancy among post-EDFA powers. Additionally, a novel method for autonomous adjustment of the operating point of amplifiers in an EDFA cascade by using a multilayer perceptron neural network is presented in \cite{barboza2013self}. The aim of this adjustment is to optimize the performance of the link by minimizing both the noise figure and the ripple of the frequency response of the transmission system while ensuring predefined input and output power levels. 
	
	\subsection{Performance Monitoring}
	
	A challenge in network control and management is to adapt to the time-varying link performance parameters, such as optical signal to noise ratio (OSNR), nonlinearity factors, chromatic dispersion (CD) and polarization mode dispersion (PMD). This subsection analyzes the suitability of the application of AI techniques in monitoring some of the aforementioned factors.
	
	The estimation and acquisition of physical parameters of transmitted optical signals allow network-diagnosis in order to take actions (repairing damages, driving compensators/equalizers or rerouting traffic around non-optimal links) against malfunctions \cite{dong2016optical}. As an example, Wu \textit{et al}.\ \cite{wu2009applications} present an extensive study of the application of artificial neural networks in optical performance monitoring (OPM), which includes the simultaneous identification of accumulated nonlinearity, OSNR, CD and PMD, from eye-diagram and eye-histogram parameters, while Szafraniec \textit{et al}.\ \cite{szafraniec2013performance} propose Kalman filter as an estimator for carrier phase tracking, polarization tracking, and estimation of the first-order PMD. However, techniques applied in \cite{wu2009applications} and similar ones \cite{li2010signed,khan2010osnr} require prior knowledge about the type of signal (bit-rate and modulation format), or additional cross-layer communication is required at the intermediate nodes to acquire this information from the upper-layer protocols, which would result in a significant increase in node complexity. In this context, a novel technique for simultaneous linear impairments identification (OSNR, CD and PMD) that is independent from bitrate and modulation format, provided that this information belongs to a known set, is proposed in \cite{tan2014simultaneous}. The study is performed using principal component analysis-based pattern recognition on asynchronous delay-tap plots and it yields accurate results in the simultaneous monitoring of linear impairments.
	
	Another recent work facing the limited scalability of the studies previously mentioned, which are based on the prior knowledge of a determined set of signals is presented in \cite{tanimura2016osnr}, where a deep neural network (DNN), trained with raw data asynchronously sampled by a coherent receiver is proposed for OSNR monitoring. Results show that OSNR is accurately estimated. Yet, this DNN needs to be configured with at least 5 layers and needs to be trained with 400,000 samples to achieve accurate results, requiring long training time. Alternately, Thrane \textit{et al}.\ \cite{thrane2017machine} propose an OSNR estimator and a modulation format classifier for systems employing advanced modulation formats (up to 64 QAM) and direct detection. The OSNR estimator employs a neural network, while the modulation format classifier uses a support vector machine (SVM), both in order to learn a continuous mapping function between input features extracted from the power eye-diagram after the photodetector and the reference OSNR and modulation format, respectively. Although accurate results are obtained for OSNR estimation and modulation format classification, the study only considers white Gaussian noise, while ignoring for the moment linear and nonlinear optical fiber impairments.
	
	\subsection{Receivers and Mitigation of Nonlinearities}
	
	Currently, the information capacity of fiber optic systems is limited by nonlinear effects of the optical fiber. Extensive research effort has attempted to address mitigation of nonlinearities on the transmission over optical fiber. Among these nonlinearities, nonlinear phase noise (NLPN) is one of the prominent factors. So far this issue has been treated with electronic methods relying on the deterministic information of the fixed fiber link, like maximum likelihood estimation \cite{lau2007signal}, digital back propagation \cite{napoli2014reduced} and stochastic digital back propagation \cite{irukulapati2016stochastic}, which may be computationally too heavy for practical implementation. 
	
	Currently, machine learning techniques are being incorporated to digital signal processing to mitigate nonlinearities in a more efficient way, allowing more accurate symbol detection. As an example, a cognitive digital receiver is proposed in \cite{gonzalez2010cognitive}, which, by means of clustering algorithms, is able to identify the incoming signal format, QPSK/8PSK/16QAM, without the need to receive a prior control message, thus opening the door to the autonomous modification of the modulation format. In addition, state-space models in combination with Bayesian filtering and expectation maximization are presented in \cite{zibar2016machine} with the aim of taking into account the underlying physics of the channel and optical elements in the formulation of signal processing algorithms. As a result, an overall system improvement is achieved, including cross-polarization mitigation, carrier synchronization and optimal symbol detection. However, expectation maximization depends on the parameters of the transmission link and consequently it is not applicable to dynamic optical networks. 
	
	Furthermore, Wang \textit{et al}.\ \cite{wang2015nonlinear} propose a machine learning algorithm to mitigate NLPN affecting M-ary phase-shift keying (M-PSK) based coherent optical transmission systems. Specifically, the algorithm introduced is a nonlinear SVM classifier able to generate nonlinear decision boundaries that allows to bypass the errors induced by nonlinear impairments in the constellations of M-PSK signals, resulting in improvements both in the maximum transmission distance and launch power dynamic range. Notwithstanding, SVM is basically a binary classifier, so to deal with higher order modulation formats, many SVMs would be necessary.
	
	Drawbacks derived from both previously mentioned studies are solved in \cite{wang2016knn}, where a k-nearest neighbors-based detector is described and demonstrated. This algorithm only needs a small set of labeled data in order to learn the link properties and generate the nonlinear decision boundaries. Moreover, it performs a multi-class classification and, therefore, it is capable of classifying multiple kinds of data simultaneously. In this way, maximum transmission distance and nonlinear tolerance improvements are demonstrated in a 16 QAM coherent transmission system. Following the same line of study, Torres \textit{et al}.\ \cite{torres2016adaptive} propose a non-symmetric demodulation technique for receivers equipped with DSP based on clustering (using k-means algorithm), which mitigates the effect of time-varying impairments such as imbalance of in-phase and quadrature signals (IQ imbalance), bias drift and phase noise. This machine learning-based demodulator is computationally highly efficient and also transparent with respect to the nonlinearity source. Finally, a recent study \cite{giacoumidis2017nonlinear} extends previous approaches by introducing these techniques in more advanced systems, with greater spectral efficiency, such as coherent optical orthogonal frequency division multiplexing (CO-OFDM) systems. The proposed algorithm is a nonlinear equalizer SVM of reduced classifier complexity using the Newton-method (N-SVM). It achieves an effective handling of inter-subcarrier nonlinear crosstalk effects and an increase of the launched optical power with low computational load.
	
	\subsection{Quality of Transmission (QoT) Estimation}
	
	Optical connection (or lightpath) QoT estimation prior to deployment is particularly relevant in impairment-aware optical network design and operation. Azodolmolky \textit{et al}.\ \cite{azodolmolky2011experimental} presented a QoT estimator tool, the Q-Tool, which computes the associated Q-factors of a set of lightpaths, given a reference topology, by combining analytical models and numerical methods. These estimates are relatively accurate, but the necessary high computing time to perform the calculations makes this tool impractical in scenarios where time constraints are important. Several approaches propose cognitive techniques to solve this drawback. As an example, Jim\'enez \textit{et al}.\ \cite{jimenez2012cognitive} present a QoT estimator capable of exploiting previous experience and thus, provide with fast and correct decisions on whether a lightpath fulfils QoT requirements or not. It is based on case-based reasoning (CBR) \cite{aamodt1994case}, an artificial intelligence mechanism that offers solutions to new problems by retrieving the most similar cases faced in the past whether by reusing them or after adapting them. Cases are retrieved from a knowledge base (KB), which can be static \cite{jimenez2012cognitive} or optimized with learning and forgetting techniques \cite{jimenez2013cognitive}. Results for CBR relying on an optimized KB show an excellent rate of successful classification of lightpaths into high/low QoT categories and more important, up to four orders of magnitude faster than the Q-Tool mentioned above. Furthermore, this study is experimentally demonstrated in a WDM 80 Gb/s PDM-QPSK testbed \cite{caballero2012experimental}, where, even with a very small KB, very high rates of successful classifications of lightpaths are achieved. One step further, and with the aim of further reducing the prediction time, Mata \textit{et al}.\ \cite{Mata2017SVM} propose the use of an SVM classifier to predict if a lightpath fulfils QoT requirements or not. This classifier proves to be not only significantly faster but also more accurate than the proposal in \cite{jimenez2013cognitive}.
	
	Another proposal for QoT estimation is that of Barletta \textit{et al}.\ \cite{barletta2017qot}, who apply a machine learning-based classifier, specifically a random forest, to predict the probability that the BER of a candidate lightpath will not exceed a given threshold. Finally, Oda \textit{et al}.\ \cite{oda2017learning} present the concept of ``living network", an optical network which keeps records of its path-level performance, which takes advantage of BER information monitoring and of a learning process (based on linear regression) in order to estimate the BER of each new service request.
		
	\section{Applications of AI in Optical Networking} \label{sec:role-AI-networks}
	
	\begin{table*}[ht!]
		\centering
		\scriptsize
		\caption{Applications in optical networking taking advantage of AI techniques (I)}
		\label{table:AI-networks-I}
		\begin{tabular}{@{}lll@{}}
			\toprule
			\midrule
			\multicolumn{1}{l}{\textbf{Applications} }                                                     & \multicolumn{1}{l}{\textbf{AI techniques} }& \multicolumn{1}{c}{\textbf{Literature}                                                                                                                                                                                                                         } \\ \midrule 
			\multirow{2}{*}{Survivable optical  networks}                                                 & Genetic algorithms                 & \begin{tabular}[c]{@{}l@{}} \cite{morais2011genetic}:\,\,\, designs the physical topology of the network ensuring survivability. \end{tabular}  \\& Ant colony optimization  & \begin{tabular}[c]{@{}l@{}} \cite{Morgan2015Ant}: \,\,\,\,\,\,guarantees survivability of the underlying network whilst minimizing the number of regenerators.\end{tabular}                                                                                                                                            
			\\ \midrule
			\multirow{4}{*}{Regenerator placement}                                                        & Genetic algorithms               & \begin{tabular}[c]{@{}l@{}} \cite{martinelli2014genetic}:\,\,\,\,\,\, minimizes the number of all-optical regenerators in order to minimize network cost.\end{tabular} \\  & & \begin{tabular}[c]{@{}l@{}} \cite{Cerutti2014Trading}: \,\,\,optimizes the selection of regenerators, code rate, and routing and spectrum allocation in flex-grid\\ \,\,\,\,\,\,\,\,\,\,\,\,\,\,\,  code-rate adaptive optical networks.                                                                                                                                                 \end{tabular} \\& Ant colony optimization  & \begin{tabular}[c]{@{}l@{}} \cite{Morgan2015Ant}: \,\,\,\,\,\,guarantees survivability of the underlying network whilst minimizing the number of regenerators.\end{tabular}                   
			\\ \midrule
			\multirow{7}{*}{Resource allocation}                                                        & Genetic algorithms                & \begin{tabular}[c]{@{}l@{}}  \cite{Velasco2013Saving}:\,\,\, proposes a novel network architecture interconnecting a set of IP/MPLS areas, performing routing \\ \,\,\,\,\,\,\,\,\,\,\,\,\,\,\,\, and flow aggregation, through a flex-grid optical core. \end{tabular}\\& & \begin{tabular}[c]{@{}l@{}} \cite{demiguel2009genetic}:\,\,\, proposes a method for joint routing and dimensioning of dynamic WDM ring networks.\end{tabular} \\& Particle swarm optimization  & \begin{tabular}[c]{@{}l@{}} \cite{Durand2013Energy}: \,\,\,solves the problem of resource allocation under Quality of Service (QoS) restrictions and the energy\\ \,\,\,\,\,\,\,\,\,\,\,\,\,\,\,\,  efficiency constraint problem.\end{tabular}                                                                                                                                             \\& Ant colony optimization                & \begin{tabular}[c]{@{}l@{}}\cite{Paula2016WDM}: \,\,\,allocation of transmitted power for energy efficient optical WDM/OCDM networks.\end{tabular}
			\\& K-means clustering                & \begin{tabular}[c]{@{}l@{}}\cite{Petridou2008Clustering}:\,\,\, proposes a message scheduling algorithm that addresses both the message sequencing and channel \\ \,\,\,\,\,\,\,\,\,\,\,\,\,\,\,\, assignment issues for a WDM star network.\end{tabular}
			\\&Markov decision processes&
			\begin{tabular}[c]{@{}l@{}}\cite{reyes2017adaptive}: \,\,\,models the resource allocation problem as a MDP to optimize an objective arbitrarily\\ \,\,\,\,\,\,\,\,\,\,\,\,\,\,\, defined by the network operator.\end{tabular}
			
			\\ \midrule
			\multirow{24}{*}{Connection establishment}                                         &Swarm intelligence
			& \begin{tabular}[c]{@{}l@{}}\cite{rubio2012comparative}:  \,\,\,proposes multiobjective algorithms based on swarm intelligence to solve the RWA problem.\end{tabular} \\&Genetic algorithms                & \begin{tabular}[c]{@{}l@{}} \cite{martinelli2014genetic}: \,\,\,\,\,\,solves the RWA problem while also ensuring QoT of lightpaths to be established.\end{tabular} \\& & \begin{tabular}[c]{@{}l@{}}\cite{Monoyios2011Multiobjective}: \,\,\,\,\,\,solves the Impairment Aware static RWA problem.\end{tabular} \\& & \begin{tabular}[c]{@{}l@{}}\cite{gong2012two-population}: \,\,\,solves the RMLSA problem.\end{tabular} \\& & \begin{tabular}[c]{@{}l@{}} \cite{lechowicz2016genetic}: \,\,\,considers multicast flows for RSA using different selection and crossover strategies.  \end{tabular}\\& &\begin{tabular}[c]{@{}l@{}}\cite{Wright2015Minimum}: \,\,\,solves the RSA problem in flex-grid networking producing useful insights into network design.\end{tabular}
			\\ & & \begin{tabular}[c]{@{}l@{}}\cite{przewozniczek2015towards}: \,\,\,solves the RSA problem with joint anycast and unicast demands.\end{tabular} \\ &Ant colony optimization   & \begin{tabular}[c]{@{}l@{}}\cite{Wang2014Distributed}: \,\,\,\,\,\,RWA solution with great robustness and adaptability to varying network and traffic conditions.\end{tabular} \\& & \begin{tabular}[c]{@{}l@{}} \cite{kyriakopoulos2016energy}: \,\,\,reduces network's energy footprint by finding the most energy-efficient routes.\end{tabular} \\& & \begin{tabular}[c]{@{}l@{}} \cite{Yuanyuan2016Optimized}: \,\,\,introduces a heuristic on the way ants choose a request from demand space in order to find the \\ \,\,\,\,\,\,\,\,\,\,\,\,\,\,\,\,shortest path. \end{tabular} \\ &Case-based reasoning   & \begin{tabular}[c]{@{}l@{}}\cite{Chen2014Cognitive}: \,\,\,reduces computing complexity when solving the traditional RWA problem in dynamic WRONs. \end{tabular} \\  
			& Simulated annealing & \begin{tabular}[c]{@{}l@{}}\cite{christodoulopoulos2011elastic}: \,\,\,solves the RMLSA problem in elastic networks.\end{tabular} \\
			& & \begin{tabular}[c]{@{}l@{}} \cite{perello2016flex}: \,\,\,solves the RMCSA problem in elastic networks with space division multiplexing (SDM). \end{tabular} \\  
			& & \begin{tabular}[c]{@{}l@{}} \cite{aibin2014simulated}: solves the RSA problem with joint anycast and unicast demands. \end{tabular} \\  
			
			& Tabu search                & \begin{tabular}[c]{@{}l@{}}\cite{walkowiak2014routing}, \cite{goscien2015tabu}: solve RSA problem providing also dedicated path protection.  \end{tabular} \\
			
			&  \begin{tabular}[c]{@{}l@{}}Backpropagation neural \\ network \end{tabular} & \begin{tabular}[c]{@{}l@{}}\cite{Jia2016Efficient}: \,\,\,predicts the arrival time and holding time of future connections by considering past experiences.\end{tabular} \\    
			& & \begin{tabular}[c]{@{}l@{}}\cite{xiong2017power}: proposes neural network-based traffic prediction to eliminate unnecessary lightpath\\ \,\,\,\,\,\,\,\,\,\,\,\,\,\,\,\,\,termination and re-establishment operations.\end{tabular} \\ 
			
			&  Q-learning  & \begin{tabular}[c]{@{}l@{}}\cite{kiran2007reinforcement}: \,\,\,solves path and wavelength selection in OBS networks.\end{tabular} 
			\\    &  Game theory  & \begin{tabular}[c]{@{}l@{}}\cite{zhu2017leveraging}: \,\,\,solves RSA problem by properly balancing the spectrum utilization and security-level of the \end{tabular}
			\\ & & \begin{tabular}[c]{@{}l@{}}\,\,\,\,\,\,\,\,\,\,\,\,\,\,\,\,\,domain in multidomain EONs.\end{tabular}\\& Neural networks and & \begin{tabular}[c]{@{}l@{}}\cite{Araujo2015Methodology}: \,\,\,estimates the connection blocking probability.\end{tabular}
			\\ & principal component analysis& \\&
			Kalman filters &\begin{tabular}[c]{@{}l@{}}\cite{yannuzzi2016interdomain}: \,\,\,reduces blocking ratio by estimating the wavelength occupancy prior to the RWA decision.\end{tabular}
			
			\\&
			Markov decision processes &\begin{tabular}[c]{@{}l@{}}\cite{tachibana2010mdp-based}: \,\,\,derives the optimal lightpath establishment policy for each service class by means of a MDP.\end{tabular}
			\\& & \begin{tabular}[c]{@{}l@{}}\cite{sue2011dynamic}: \,\,\,proposes the use of an efficient dynamic-preemption call admission control scheme\end{tabular}\\& & \begin{tabular}[c]{@{}l@{}}\,\,\,\,\,\,\,\,\,\,\,\,\,\,\,\,\,based on the optimal policy derived from a MDP.\end{tabular}
			
			\\ \midrule
			
			\multirow{8}{*}{Network reconfiguration} &\begin{tabular}[c]{@{}l@{}}Genetic algorithms and\\ ant colony optimization\end{tabular} & \begin{tabular}[c]{@{}l@{}} \cite{Corut2010Ensuring}: \,\,\,survivable mapping of a given WDM virtual topology. \end{tabular}      \\  &Genetic algorithms & \begin{tabular}[c]{@{}l@{}}\cite{fernandez2012survivable}: \,\,\,designs virtual topologies while reducing energy consumption and network congestion.\end{tabular} \\& & \begin{tabular}[c]{@{}l@{}} \cite{fernandez2013techno}: \,\,\,techno-economic study of the introduction of cognitive techniques in virtual topology design. \end{tabular} \\& & \begin{tabular}[c]{@{}l@{}} \cite{Gao2015Virtual}: \,\,\,addresses reliable multicast Virtual Network mapping for OFDM based EONs. \end{tabular}      \\  & \begin{tabular}[c]{@{}l@{}}Genetic algorithms and\\ cognition \end{tabular} & \begin{tabular}[c]{@{}l@{}}\cite{fernandez2015virtual}:  \,\,\,produces estimations that can help to anticipate changes in the traffic and proactively reconfigure \\ \,\,\,\,\,\,\,\,\,\,\,\,\,\,\, the virtual network topology.\end{tabular}      \\  & Neural networks & \begin{tabular}[c]{@{}l@{}}\cite{morales2017virtual}: \,\,\,performs reconfigurations based on the traffic volume and direction predicted by a neural network.\end{tabular}
			\\ & Markov decision processes &
			\begin{tabular}[c]{@{}l@{}}\cite{baldine2001traffic}: determines how frequently to reconfigure a broadcast WDM optical network based on MDPs.\end{tabular}
			\\ \midrule

			\multirow{5}{*}{Failure/fault detection} & Bayesian networks, clustering      &  \begin{tabular}[c]{@{}l@{}}\cite{ruiz2016service}: \,\,\,identifies or locates failures in the virtual network topology to improve quality of service.\end{tabular}  \\ 
			& Cognition-based methods       & \begin{tabular}[c]{@{}l@{}} \cite{Zhang2016Failure}:  \,\,\,detects failures in centralized SDN-based networks by periodically exchanging the messages \\ \,\,\,\,\,\,\,\,\,\,\,\,\,\,\, between controller and switches.\end{tabular} \\ 
			& Bayesian inference/networks &\begin{tabular}[c]{@{}l@{}} \cite{gosselin2017application}, \cite{Tembo2016Tutorial}: probabilistic modeling and machine learning for fault diagnosis in optical access networks\end{tabular} \\ \midrule
			
			\multirow{5}{*}{Software Defined Networking} & Cognition-based methods & \begin{tabular}[c]{@{}l@{}}\cite{Montero2017Dynamic}: \,\,\,correct mapping of topologies in considerably low total times.\end{tabular} \\& & \begin{tabular}[c]{@{}l@{}}  \cite{oliveira2015toward}:  introduces a transport SDN controller that facilitates optical network virtualization and\\ \,\,\,\,\,\,\,\,\,\,\,\,\,\,\,\,\,autonomic operation. \end{tabular} \\& & \begin{tabular}[c]{@{}l@{}} \cite{Yoo2014Multi}: proposes a new inter-networking paradigm based on broker agents with cognitive intelligence.\end{tabular} \\ & Neural networks &  \begin{tabular}[c]{@{}l@{}}
				\cite{Yan2017Field}: maximizes link capacity after predicting link performance in correlation with the OSNR. \end{tabular}
			\\ \midrule
			
			\multirow{12}{*}{Reduction/estimation of burst loss} &Learning automata &  \begin{tabular}[c]{@{}l@{}} \cite{praveen2006first}: achieves self-awareness, self-protection and self-optimization in OBS networks.\end{tabular} \\ & Q-learning &  \begin{tabular}[c]{@{}l@{}} \cite{kiran2007reinforcement}:  \,\,\,solves path and wavelength selection in OBS networks. \end{tabular} \\& & \begin{tabular}[c]{@{}l@{}} \cite{venkatesh2008complete}: exploits the feedback loop to control the retransmission rate of bursts that are lost.\end{tabular} \\& & \begin{tabular}[c]{@{}l@{}} \cite{Haeri2013Reinforcement}:  introduces a low-complexity solution to resolve contention in OBS networks\end{tabular} \\& \begin{tabular}[c]{@{}l@{}}Hidden Markov model\\ and expectation-maximization \end{tabular} & \begin{tabular}[c]{@{}l@{}} \cite{jayaraj2008loss}: \,\,\,proposes variations of the TCP protocols to enhance the performance of OBS networks.\end{tabular} \\& Bayesian networks &  \begin{tabular}[c]{@{}l@{}}\cite{Elbiaze2011Cognitive}: decreases the burst loss ratio (BLR) in a OBS network.\end{tabular} \\&\begin{tabular}[c]{@{}l@{}}   Feed-forward neural network\\ and Q-learning  \end{tabular} & \begin{tabular}[c]{@{}l@{}} \cite{Haeri2015Intelligent}: proposes deflection routing protocols that achieve smaller burst-loss probabilities \\ \,\,\,\,\,\,\,\,\,\,\,\,\,\,\,\,\,than previous approaches while deflecting bursts less frequently. \end{tabular}\\   & Extreme learning machine & \begin{tabular}[c]{@{}l@{}}\cite{Leung2017Extreme}: estimates burst loss probability. \end{tabular}\\
			&Ant colony optimization&
			\begin{tabular}[c]{@{}l@{}}\cite{coulibaly2015qos-aware}: proposes an ACO approach to reduce burst loss ratio, enhancing at the same time the delay.\end{tabular}\\
			    \midrule

			\bottomrule
		\end{tabular}
	\end{table*}
	
\begin{table*}[ht!]
	\centering
	\scriptsize
	\caption{Continuation: Applications in optical networking taking advantage of AI techniques (II)}
	\label{table:AI-networks-II}
	\begin{tabular}{@{}lll@{}}
		\toprule
		\midrule
		\multicolumn{1}{l}{\textbf{Applications} }                                                     & \multicolumn{1}{l}{\textbf{AI techniques} }& \multicolumn{1}{c}{\textbf{Literature}                                    } \\ \midrule 
		Statistical solutions for prediction                                                  & Hidden Markov model (HMM)          & \begin{tabular}[c]{@{}l@{}}\cite{chitra2014hidden}: \,\,\,uses HMM based traffic prediction along with QoS aware light path establishment in WDM\\ \,\,\,\,\,\,\,\,\,\,\,\,\,\,\,  networks.\end{tabular}\\ & Bayesian methods and game theory & \begin{tabular}[c]{@{}l@{}}\cite{awwad2009bayesian}: \,\,\,utilizes a Bayesian game-theoretic model to  guarantee cooperativeness in RF/FSO networks.\end{tabular} \\ \midrule
		Intelligent ROADM                                                                     & Linear regression  &            \begin{tabular}[c]{@{}l@{}}\cite{oda2017learning}: \,\,\,autonomously keeps  record of  path-level performance. \end{tabular}\\ \midrule
		Splitter placement in PONs & Genetic algorithms & \cite{Villalba2009Design}, \cite{Kokangul2011Optimization}, \cite{Liu2012Nested}: these papers optimize the location of splitter to achieve various objectives in PON. \\ \midrule 
		
		\multirow{2}{*}{
			\begin{tabular}[c]{@{}l@{}}QoS guarantees and dynamic\\ bandwidth allocation in PONs\end{tabular}
		} & \begin{tabular}[c]{@{}l@{}}Genetic algorithms and\\ neural networks\end{tabular} & \cite{Huang2008Study}, \cite{Moradpoor2013Mathematical}, \cite{Jimenez2015Auto},\cite{Hwang2012Genetic}, \cite{Merayo2017PID}: use genetic algorithms or neural networks to assure QoS in PON
		\\ & Bayesian estimation & \cite{Dias2014Bayesian}: proposes estimation and prediction-based Just-In-Time dynamic bandwidth allocation algorithm. \\ \midrule 
				
		\multirow{2}{*}{Placement of ONUs} &Teaching learning-based optimization & \begin{tabular}[c]{@{}l@{}}\cite{Bhatt2017ONU}:  reduces the required Optical Network Units (ONUs) that assure connectivity among \\ \,\,\,\,\,\,\,\,\,\,\,\,\,\, wireless routers and ONUs in a Fiber-Wireless network. \end{tabular} \\ & Genetic algorithms & \begin{tabular}[c]{@{}l@{}}\cite{Bhatt2015Hybrid}: another non-deterministic approach for placement of ONUs. \end{tabular} \\ \midrule 
		\begin{tabular}[c]{@{}l@{}}Cognitive optical networks \end{tabular} &  Cognition-based methods                                 & \begin{tabular}[c]{@{}l@{}}\cite{de2013cognitive, borkowski2015cognitive, zervas2010cognitive, wei2012cognitive, yoo2014multi-domain, chan2017cognitive}: these papers propose cognitive optical network architectures. \end{tabular}                                                                                                                               \\ \midrule 
		\multirow{4}{*}{Intra-Datacenters} & Multilayer perceptron & \begin{tabular}[c]{@{}l@{}} \cite{Rastegarfar2016TCP}: \,\,\,allocates resources (optical circuits/electrical switches) to flows according to their requirements.\end{tabular} \\ & Neural networks & \begin{tabular}[c]{@{}l@{}}\cite{Glick2017Scheduling}: presents a flow classifier at the edge of the network combined with an SDN centralized controller.\end{tabular} \\ & Markov decision processes & \begin{tabular}[c]{@{}l@{}}\cite{Wang2017Adaptive}: makes scheduling decisions in all-optical data center networks guaranteeing throughput  optimality\\ \,\,\,\,\,\,\,\,\,\,\,\,\,\,\, under a zero reconfiguration delay. \end{tabular} \\ \midrule 
		
		\bottomrule
	\end{tabular}
\end{table*}

	AI presents several opportunities for automating operations and introducing intelligent decision making in network planning and in dynamic control and management of network resources, including issues like connection establishment, self-configuration and self-optimization, through prediction and estimation by utilizing present network state and historical data. In this section, we review these applications as well as use cases of AI in optical burst-switched networks (OBS), in passive optical networks (PONs) and intra-datacenter networks. These applications are summarized in Tables \ref{table:AI-networks-I} and \ref{table:AI-networks-II}.

	\subsection{Optical Network Planning} \label{subsec:planning}
	
	As described in Section \ref{sec:AI-techniques}, search algorithms and optimization theory have been widely used for optical network planning and dimensioning (e.g., \cite{mukherjee2006optical, simmons2014optical}), usually complemented or extended with local search algorithms and metaheuristics like simulated annealing, swarm optimization and genetic algorithms \cite{du2016search}.
	
	Optical network planning involves tasks like designing the physical topology of the network and ensuring survivability while minimizing costs. Morais \textit{et al}.\ \cite{morais2011genetic} propose the use of genetic algorithms (GAs) to address those issues in an opaque optical transport network, and de Miguel \textit{et al}.\ \cite{demiguel2009genetic} also rely on a GA for dimensioning dynamic WDM ring networks. A related optimization problem, like minimizing the number of all-optical regenerators, is tackled by Martinelli \textit{et al}.\ \cite{martinelli2014genetic} with a GA, which also jointly solves the routing and wavelength assignment (RWA) problem while ensuring the QoT for the lightpaths to be established. The problem of placing regenerators in optical networks subject to fault tolerance constraints has also been approached by means of ant colony optimization (ACO) techniques \cite{Morgan2015Ant}. This proposal guarantees survivability of the underlying network whilst also minimizing the number of regenerators required. High quality solutions are provided, with reasonable runtimes. Furthermore, a GA aiming to jointly optimize the selection of nodes performing 3R regeneration, code rate, and routing and spectrum allocation for lightpaths to be established in flex-grid code-rate adaptive optical networks is proposed in \cite{Cerutti2014Trading}. The addition of the code rate in the classical resource and allocation problem entails establishing a trade-off between minimizing the number of regeneration nodes and minimizing the need of spectral resources. Results indicate that, in general, with just few nodes selected for regeneration, it is possible to ensure QoT and exploit the advantages of code-rate adaptiveness. In the context of flex-grid network planning, it is worthy to note the work by Velasco \textit{et al}.\ \cite{Velasco2013Saving}. They propose a novel network architecture consisting of a set of IP/MPLS areas performing routing and flow aggregation, which are interconnected through a flex-grid optical core. In order to obtain near-optimal solutions for this architecture for real-sized network and traffic instances, they employ GAs (in particular, biased random-key GAs). Under these circumstances, simulation results reveal that extending the core toward the edges results in significant savings in both the core and IP/MPLS networks. 
	
	A different approach, \cite{Durand2013Energy}, presents a particle swarm optimization (PSO) algorithm in order to solve the problem of resource allocation based on the signal-to-noise plus interference ratio optimization in a hybrid wavelength division multiplexing/optical code division multiplexing network (WDM/OCDM) under Quality of Service (QoS) restrictions and the energy efficiency constraint problem. The PSO strategy allows the regulation of the transmitted power in order to maximize the energy efficiency. Results show interesting trade-offs between performance and complexity. Following the same trend, \cite{Paula2016WDM} presents an alternative algorithm, a heuristic ACO scheme, for allocation of transmitted power with increasing energy efficiency applicable to optical WDM/OCDM transport networks. In addition, an analytical disciplined convex optimization approach, taking into account the performance and complexity metrics, is proposed as comparison. Simulation results demonstrate that the ACO scheme proves to be useful in order to obtain spectral-efficient and energy efficient systems suitable for WDM/OCDM networks, with promising performance\textendash complexity trade-offs in comparison with the analytical approach.
	
	Another example of the use of AI in resource allocation is the work by Petridou \textit{et al}.\ \cite{Petridou2008Clustering}. They propose a message scheduling algorithm, based on the k-means clustering algorithm, which addresses both message sequencing and channel assignment for a WDM star network. Based on the produced clusters, the scheduling algorithm manages to avoid scheduling consecutive messages to the same destination which harms the channels\textquoteright{} utilization.

	\subsection{Connection Establishment} \label{subsec:connection}
	
	Metaheuristics like simulated annealing and evolutionary methods like genetic algorithms or particle swarm optimization, are effective in solving hard optimization problems because they are less likely to become trapped in local optima. Therefore, these methods are useful to solve the optical connection (lightpath) establishment problem in optical networks. In WDM networks, this involves searching a combination of route and available wavelength, and is so called the routing and wavelength assignment (RWA) problem. In elastic optical networks (EONs), it involves searching for a route and a portion of available spectrum and even a modulation format, i.e., solving the routing and spectrum allocation (RSA) or the routing, modulation level and spectrum allocation (RMLSA) problems.
	
	A multi-objective GA for solving the impairment-aware static RWA problem is presented in \cite{Monoyios2011Multiobjective}, and Rubio-Largo \textit{et al}.\  \cite{rubio2012comparative} present a comparative study among three multiobjective evolutionary algorithms (MOEAs) based on swarm intelligence to solve the RWA problem in real-world optical networks: artificial bee colony algorithm, gravitational search algorithm and firefly algorithm, concluding that swarm intelligence is very suitable for this task. 
	
	Wang \textit{et al}.\ \cite{Wang2014Distributed} include considerations of mixed line rate, physical impairments and traffic grooming functionality to solve the RWA problem by means of an ACO algorithm. Different configurations of this distributed solution are compared to each other and also with a centralized grooming adaptive shortest path algorithm. Although the centralized solution shows better efficiency in terms of blocking probability, ACO shows great robustness and adaptability to varying network and traffic conditions. Additionally, in \cite{Chen2014Cognitive}, a cognitive approach (case-based reasoning) is introduced into the traditional RWA algorithm for dynamic WRONs with the aim of reducing computing complexity. Simulation results indicate that taking advantage of similar past experiences or cases stored in a knowledge base (KB) can reduce computational time by 25\% over classical RWA algorithms, while maintaining or even improving performance. In addition, Kyriakopoulos \textit{et al}.\ \cite{kyriakopoulos2016energy} propose a heuristic method based on ACO to reduce network energy footprint by exploiting the basic principles of swarm intelligence for finding the most energy-efficient routes from source to the destination node per traffic request. A different ACO-based proposal \cite{Yuanyuan2016Optimized}, which introduces a heuristic on the way ants choose a request from demand space (those that can be served with shorter route first), outperforms both regular ACO and shortest-path and most-used algorithms. Additionally, Ara\'{u}jo \textit{et al}.\ \cite{Araujo2015Methodology} present a mechanism to estimate the blocking probability when establishing lightpaths in an optical network. It consists of an artificial neural network which uses as inputs topological properties and general physical layer characteristics (on which a principal component analysis is previously carried out). Results show a speed-up greater than 7500 times than that of a discrete event simulator, and accurate estimates are obtained except when very small blocking probabilities are evaluated.
	
	The RSA problem in EONs is NP-hard \cite{Klinkowski2011Routing}, and has also been considered an appropriate candidate to be solved by metaheuristics. In fact, in the seminal paper by Christodoulopoulos \textit{et al}.\ \cite{christodoulopoulos2011elastic} on elastic bandwidth allocation, simulated annealing is used to solve the RMLSA problem. Simulated annealing has been also used by Perell\'o \textit{et al}.\ \cite{perello2016flex} in elastic networks with space division multiplexing (SDM), where not only route, modulation format, and spectrum have to be assigned but also a fiber core, finally solving the route, modulation format, core, and spectrum assignment (RMCSA). 
	
	In the context of evolutionary methods, the RMLSA problem has been solved by means of a GA where two populations evolve in parallel and use a migration operation to exchange individuals between them \cite{gong2012two-population}. Furthermore, the RSA problem has also been solved in \cite{lechowicz2016genetic} with GAs for multicast flows considering different selection and crossover strategies, and in \cite{walkowiak2014routing} with tabu search techniques with the aim of providing dedicated path protection. Another coevolutive approach is introduced in \cite{przewozniczek2015towards} to solve the RSA problem with joint anycast and unicast demands, outperforming previous proposals based on tabu search \cite{goscien2015tabu} and simulated annealing \cite{aibin2014simulated} approaches.
	
	An alternative approach to solve the RSA problem in EONs is presented in \cite{Jia2016Efficient}, where a backpropagation neural network is proposed to improve the RSA algorithm by predicting the arrival time and holding time of future connections by considering past experiences. Results confirm this approach outperforms RSA algorithms that do not make use of historical information. A backpropagation neural network is also used for traffic prediction by Xiang \textit{et al}.\ \cite{xiong2017power}. They propose this technique in the context of centralized lightpath management in inter-datacenter EONs, with the aim of reducing switching power consumption by eliminating unnecessary lightpath termination and re-establishment operations. Another work, \cite{Wright2015Minimum}, applies a Shannon entropy-based fragmentation metric to the RSA problem in flex-grid networking by utilizing two complementary approaches: minimum and maximum entropy. The former allows to increase the number of demands that can be served before reaching critical blocking levels by reducing as much as possible spectrum fragmentation. The latter, where source-destination pair bandwidth demands are located as far apart from one another as possible across the optical spectrum solved, employs a GA-based optimization in order to produce useful insights into network design.

	\subsection{Network Reconfiguration: Virtual Topologies} \label{subsec:reconfiguration}
	
	The virtual topology is the set of optical connections (or lightpaths) established in a network. It does not have to be statically configured, but it can be dynamically reconfigured in order to better adapt to evolving traffic demands with some objectives like reducing energy consumption, network congestion, end-to-end delay or blocking probability or trying to ensure quality of transmission (QoT), etc. For that purpose, two nature inspired heuristics, GA and ACO, are used in \cite{Corut2010Ensuring} to obtain a survivable mapping of a given WDM virtual topology. Feasible solutions are obtained even for large topologies when integer linear programming methods cannot. Also, a multiobjective genetic algorithm to design virtual topologies with the aim of reducing both the energy consumption and the network congestion is presented by Fern\'{a}ndez \textit{et al}.\ \cite{fernandez2012survivable}. The GA proposed there is enhanced with the capability of remembering solutions successfully used in the past, as well as connections with low QoT. The incorporation of those mechanisms leads to improvements in performance. Furthermore, the introduction of cognitive techniques in virtual topology design also exhibits significant savings in terms of the total cost of ownership compared to conventional methods. As a matter of fact, savings up to 20\% and 25\% in capital and operational expenditures, respectively, via a GA-based method, are demonstrated in \cite{fernandez2013techno}. One step forward, the extension presented in \cite{fernandez2015virtual} uses monitored data to produce estimations that can help to anticipate changes in the traffic and proactively reconfigure the virtual network topology. 
	
	An algorithm to identify/locate failures in the virtual network topology that can lead to an unacceptable quality of service is proposed by Ruiz \textit{et al}.\ \cite{ruiz2016service}. They first perform the experimental characterization of several causes of failure (which is done with the help of a clustering algorithm), and then use those characterizations to train a Bayesian network (BN). This trained BN is used to localize and identify the most probable cause of failure impacting a given service. 
	
	A virtual network topology reconfiguration approach is introduced in \cite{morales2017virtual}. It performs reconfigurations based on the traffic volume and direction predicted by an artificial neural network proposed for every origin-destination pair. Periodically, collected monitoring data are transformed into modelled data and the artificial neural networks are used to predict the next-period traffic. Results show savings in both capital and operational expenditures.
	
	A different approach is followed in \cite{Gao2015Virtual}, where an efficient virtual network (VN) mapping for multicast services over both general IP networks and orthogonal frequency division multiplexing (OFDM)-based EONs, is presented. This proposal takes into consideration the max-min fairness in terms of reliability among distinct VNs. In the IP networks case and with the aim to globally optimize the reliability and fairness of all the multicast VN requests, a mixed integer linear programming (MILP) model to determine the upper bound on the reliability is presented, as well as a GA that addresses reliable multicast VN mapping. For OFDM based EONs, this solution is extended by considering the most efficient modulation format selection strategy, spectrum continuity, and conflict constraints.

	\subsection{Software Defined Networking}
	
	The software defined networking (SDN) paradigm \cite{Xia2015Survey}, which decouples control and data planes, and enables programmability on the former plane, has aroused the interest of both industry and research communities by allowing networks managers to manage, configure, automate and optimize network resources via software. In the context of SDN over optical networks, a correct mapping of the underlying topology at the control plane level is crucial. Following this requirement, a novel SDN-based cost-effective topology discovery method, allowing transparent optical networks to automatically learn physical adjacencies between optical devices, is introduced in \cite{Montero2017Dynamic}. This is achieved by means of a test-signal mechanism --by exchanging and verifying identifier information between discovery agents-- and the OpenFlow protocol, resulting in correct mapping of the topologies in low total times. In relation to this paradigm, fault tolerance is of paramount importance when it comes to characterizing optical networks. In \cite{Zhang2016Failure}, an efficient cognitive process for failure detection in centralized SDNs is proposed. Specifically, a network controller interacts with optical cross-connects (OXCs) exchanging messages periodically to efficiently detect failures by using the link layer discovery protocol (LLDP). Fast and accurate communication of these link events to the controller allows a dynamic routing algorithm to update the topology and restore the optical path in a significantly short space of time.
	
	Furthermore, Oliveira \textit{et al}.\ \cite{oliveira2015toward} introduce a transport SDN controller that facilitates optical network virtualization and autonomic/cognitive operation by means of two adaptive algorithms that allow to reconfigure, on one hand, the transmission modulation format and spectrum utilization according to network conditions and on the other hand, the attenuation applied at the ROADMs to improve the OSNR of the signals at the reception. Yan \textit{et al}.\ \cite{Yan2017Field} propose and demonstrate in a field trial the planning of an SDN-based optical network utilizing neural network-based methods, which are able to predict link performance in correlation with the OSNR. By means of probabilistic-shaping bandwidth variable transmitters (BVTs), which are configured by the SDN controller based on these predictions, spectral efficiency can be adapted, maximizing the link capacity. Additionally, a new inter-networking paradigm based on broker agents with cognitive intelligence that compete to provide desirable inter-networking services to autonomous systems through market-driven incentives, is proposed in \cite{Yoo2014Multi}.

	\subsection{Applications in Optical Burst Switching}
	
	Optical burst-switched (OBS) networks \cite[Chap.~18]{mukherjee2006optical} have also taken advantage of artificial intelligence, and in particular, of machine learning techniques. Praveen \textit{et al}.\ \cite{praveen2006first} propose a novel OBS architecture which takes advantages of learning automata to achieve self-awareness, self-protection and self-optimization, consequently reducing burst loss probability significantly. Work done in \cite{praveen2006first} has been extended in different studies by using other machine learning techniques, such as Q-learning, in order to solve the path and wavelength selection problem \cite{kiran2007reinforcement}, or by exploiting the feedback loop to control the retransmission rate of bursts that are lost \cite{venkatesh2008complete}. Moreover, variations of the TCP protocols to enhance the performance of OBS networks, including supervised and unsupervised learning techniques, are also proposed in \cite{jayaraj2008loss}. 
	
	Burst blocking or loss probability --the ratio of the number of lost bursts to the total number of transmitted bursts-- is commonly used for the performance measurement of OBS network technologies. So far, proposed techniques have proven to be too slow or not accurate enough to estimate this parameter. New approaches based on machine learning have outperformed previous studies, especially in terms of computation time. Leung \textit{et al}.\ \cite{Leung2017Extreme} present two models for burst loss ratio (BLR) estimation employing neural networks based on the extreme learning machine (ELM) framework. By using these models, estimates can be obtained much faster than by means of simulations. Moreover, the accuracy of the BLR estimates outperforms that obtained with an existing analytical approach and are very close to the values obtained by simulation. Three cognitive mechanisms --Bayesian networks, closed loop control and open loop control-- to decrease BLR in an OBS network are introduced in \cite{Elbiaze2011Cognitive}. Simulation results confirm that the application of these methods in the admission process leads to a BLR reduction in OBS networks. Similarly, in \cite{Haeri2013Reinforcement} a novel node degree dependent signalling algorithm in combination with Q-learning is proposed as a low-complexity deflection routing protocol with the aim to resolve contention in OBS networks. This solution scales well for large networks, since its complexity depends on the node degree rather than network size. Simulation results show that despite its lower complexity, burst loss probability of the proposed algorithm is comparable to other existing reinforcement learning-based deflection routing algorithms. The previous work is extended in \cite{Haeri2015Intelligent}, where a framework that adds intelligence to deflection routing in buffer-less architectures is presented. In particular, by means of the combination of the node degree-dependent signalling algorithm with a feed-forward neural network (NN) and a feed-forward NN with episodic updates, both containing Q-learning-based decision-making modules, the proposed deflection routing protocols are proved to achieve smaller burst-loss probabilities than previous approaches while deflecting bursts less frequently. In addition, the solution requires less memory and CPU resources, which are more significant as the size of the network grows. Additionaly, Coulibaly \textit{et al}.\ \cite{coulibaly2015qos-aware} propose an ACO approach to reduce burst loss ratio, enhancing at the same time the delay, by means of an adaptive and quality of service (QoS)-aware route, wavelength and timeslot assignment algorithm.

	\subsection{Applications in Passive Optical Networks (PONs)}
	
	Network planning of passive optical networks (PONs) \cite[Chap.~5]{mukherjee2006optical} has also taken advantage of AI techniques. Villalba \textit{et al}.\ \cite{Villalba2009Design} propose a GA for topology searching and splitter placement in these networks. A graph representation scheme, associated with a street map, is used as a reference, and the formulation minimizes the amount of optical cabling and number of splitters and power budget. Besides, Kokangul \cite{Kokangul2011Optimization} proposes a GA and mathematical modeling techniques to optimize the position of the primary and secondary nodes (i.e., the points where the signal is split the first and the second time, respectively), their split levels, the association of customers to secondary nodes, and the association of secondary nodes to primary nodes. This placement is done under constraints such as attenuations and characteristics of the optical devices. Liu \textit{et al}.\ \cite{Liu2012Nested} propose a topology optimization model in long-reach passive optical networks (LR-PON) using a nested genetic algorithm (NGA). The outer-loop of the NGA algorithm deals with the location of splitters and the inner-loop of the algorithm builds the spanning tree. Bhatt \textit{et al}.\ \cite{Bhatt2017ONU} propose a teaching-learning based optimization (TLBO) algorithm to reduce the required optical network units (ONUs) that assure connectivity among wireless routers and ONUs in a Fiber-Wireless network. Results of the simulations carried out for different grid sizes of the geographical area and variable wireless routers confirm that the implemented scheme requires less ONUs, reaching a globally optimum solution that previous random and deterministic approaches failed to provide. Along the same line, other non-deterministic approaches for placement of ONUs also provide desired levels of performance, e.g., using genetic algorithms \cite{Bhatt2015Hybrid}.
	
	Other research works focus on the diagnosis and the self-diagnosis of PONs. As an example, Gosselin \textit{et al}.\ \cite{gosselin2017application, Tembo2016Tutorial} propose probabilistic modeling and machine learning for fault diagnosis in optical access networks. The proposal is based on a Bayesian network which encodes expert knowledge. Indeed, they develop a Bayesian inference engine, named probabilistic tool for GPON-FTTH access network self-diagnosis (PANDA), to efficiently allow fault diagnosis in GPONs. Sarigiannidis \textit{et al}.\ \cite{Sarigiannidis2017DIANA} propose a 10-gigabit-capable PON (XG-PON) together with multiple long term evolution (LTE) radio access networks in the fronthaul. The mechanism receives traffic-aware knowledge from the SDN controllers and it modifies the uplink-downlink configuration in the LTE radio communication. This strategy allows to calculate an optimal configuration based on the traffic dynamics in the global network, allowing for an improvement in the packet latency and jitter.
	
	Additional research is focused on assuring quality of service (QoS) requirements and media access control (MAC) issues in PONs. Some approaches apply GAs to deal with these issues. As an example, 
	Huang \cite{Huang2008Study} proposes to balance asymmetric traffic load between ONUs in PON architectures using a GA, which also decreases congestion in ONUs. Besides, Hwang \textit{et al}.\ \cite{Hwang2012Genetic} formulate a genetic expression programming (GEP) algorithm for the QoS traffic prediction integrated with the limited packet transmission strategy to deal with the queue variation and as a consequence with the reduction of the delay of high priority traffic. In addition, Moradpoor \textit{et al}.\ \cite{Moradpoor2013Mathematical} propose a dynamic excess bandwidth allocation algorithm for an integrated hybrid PON with wireless technology employing a GA. The algorithm is able to provide optimal/near-optimal solutions for the excess bandwidth assignment in this converged network scenario.
	
	Other approaches implement proportional-integral-derivative (PID) controllers combined with GA or ANN techniques to comply with QoS requirements. Thus, Jim\'{e}nez \textit{et al}.\ \cite{Jimenez2015Auto} develop an automatic tuning technique based on GA to tune a proportional controller to provide delay guarantees. In addition, they integrate a dynamic admission control module to transmit or drop packets in order to achieve a better delay control. Besides, same authors in \cite{Merayo2017PID} propose a PID controller integrated with an ANN to efficiently ensure QoS bandwidth requirements in EPONs. Finally, Dias \textit{et al}.\ \cite{Dias2014Bayesian} propose a dynamic bandwidth allocation algorithm which uses Bayesian estimation to estimate the average interarrival time of packets at the ONUs, with the aim of minimizing the queuing delay introduced by the sleep- and doze-mode operations.
	
	\subsection{Applications in Intra-Datacenter networking}\label{subsec:datacenters}
	
	Intra-datacenter (DC) networks are also embracing machine learning techniques in order to improve  performance. For instance, in hybrid-switching-based DCs, where an electrical packet-switched and an optical circuit-switched network live together, machine learning-based flow classification may be a decisive solution to improve speed and accuracy, besides improving adaptability to traffic dynamics. As an example, Rastegarfar \textit{et al}.\ \cite{Rastegarfar2016TCP} have proposed a multi-layer perceptron to allocate resources to TCP flows according to their requirements (e.g., allocating optical circuits to bulk data transfer and mapping short mice flows to electrical switches), and have obtained very significant improvements in network throughput due to both optical channel bandwidth consolidation and adaptive flow classification. Furthermore, a neural network flow classifier at the edge of the network, combined with an SDN centralized controller able to take advantage of this classification outcome along with its global view of the resources has been proposed in \cite{Glick2017Scheduling}. Finally, Wang and Javidi \cite{Wang2017Adaptive} have also recently considered the end-to-end scheduling problem in all-optical data centers networks, whose main challenges are the bufferless nature and the nonzero reconfiguration delay of optical switches, making it necessary to use a centralized controller that can efficiently schedule the end-to-end transmissions. The proposed method employs an adaptive Markov scheduling policy, which makes decisions every time slot, and determines both the schedule and the time to reconfigure the schedule based on the most recent queue length information.

	\section{New Opportunities and Challenges for the Use of AI in Optical Networks} \label{sec:opportunities} 
	
	In this section, we describe a number of new opportunities and challenges that we envision in the area of optical systems and networking. We envision increasingly challenging roles of the use of AI in the physical layer, where it will continue being a useful tool not only in the framework of emerging optical transmission technologies but also in helping increasing security by means of attack or intrusion detection and localization. We also describe the relevant role of AI in the automation of network management operations, and in the support of emerging networking paradigms. 
	
	\subsection{Optical Transmission Systems, and Attack and Intrusion Detection}
	
	Many control decisions in a network are made based on accurate measurement or estimation of physical parameters. Thus, in line with the advances reported in Section \ref{sec:role-AI-physical}, we expect AI will continue playing an important role in supporting emerging transmission technologies like space division multiplexing, multimode/multicore fibers and advanced modulation formats and constellation shaping. We also expect further progresses in the use of AI techniques for QoT estimation and performance monitoring. However, a key area closely related to monitoring where we see a significant opportunity for AI techniques, but have not yet emerged (to the best of our knowledge) in the optical arena, is that of attack and intrusion detection.
	
	Increasing Internet proliferation has driven businesses and other institutions to provide essential services over the Internet infrastructure. As a result, these entities have to contend with higher risk and potential costs associated with data breaches, which is currently estimated to be as high as \$3.6~million \cite{databreach-sec} on average. The optical network layer is also suceptible to attacks by malicious entities, which can be broadly classified as \emph{disruption} and \emph{eavesdropping} attacks. Disruption attacks attempt to exploit physical characteristics of optical transmission to degrade the operation of existing services and/or block incoming service requests, while eavesdropping attacks are targeted towards gaining unauthorized access to the data transmitted over the optical network. A variety of attack vectors for both these attack categories have been outlined in \cite{kapov-sec, fok-sec, medard-sec}. 
	
	Addressing attacks on optical infrastructure consists of mechanisms to detect and localize an attack on the optical infrastructure, and the mitigation mechanisms in response to a detected attack. Statistical approaches using optical measurements to identify and localize attacks in optical networks have been proposed in \cite{rejeb-sec, medard-detection} and can be employed to detect a range of attacks on the optical network infrastructure. However, the computation complexity involved in applying the proposed techniques in a large-scale network is non-trivial. Therefore, AI techniques could be applied in this context to identify and localize optical attacks. 
	
	On successful identification of attacks, mitigation techniques, especially in the context of optical telecommunication networks involves all-optical encryption or re-routing of traffic away from attack locations. The application of AI for network optimization and RWA problems has already been discussed in Sections \ref{subsec:planning} to \ref{subsec:reconfiguration}, and can be employed to move vulnerable optical traffic away form attack locations.   
	
	\subsection{Automating Network Management Operations}
	In the area of optical network operation, heterogeneous (multi-technology and multi-vendor) network devices make operation, administration and maintenance of optical networks a complex and challenging process. This is due to the fact that network state information, e.g., topology, congestion, failure discovery, etc., collected from different devices have different and limited state information, which pose a big challenge in data collection, processing and decision making. Thus, process automation has been identified as a key enabler for driving down operational costs, and also to help move human intelligence away from repetitive tasks \cite{lightreading-aut}. AI-based techniques are seen as a key enabler for automating network management operations \cite{kompella-aut, jukan2017evolution}, and cognitive optical networks were in fact proposed with that aim \cite{de2013cognitive, borkowski2015cognitive, zervas2010cognitive, wei2012cognitive, yoo2014multi-domain, chan2017cognitive} but many open issues remain. The primary pre-requisite for automating network operations is the capability to efficiently collect and analyze telemetry information from the network. SDN and related standards have introduced streaming telemetry which can be used to efficiently collect information from the network all the way to the network edge. The scale of information collected in a large telco \cite{king-aut} poses a challenge for processing and identifying anomalies such as network mis-configurations, alarm-correlation, failure localization and prediction. AI-based techniques can be efficiently used to target these problems at scale, especially when combined with Big Data frameworks like the Apache Hadoop \cite{hadoop} or Spark \cite{spark} ecosystems. Another key enabler for automating network management involves the use of \emph{declarative intents} which provide a syntax for users to describe their requirements to network orchestration platforms \cite{onos, opendaylight}. Intents are widely employed in natural language processing (NLP) and AI-based techniques from these domains can be used to integrate user interaction platforms such as chatbots, voice command devices, etc., to interpret user requests into intents for network orchestration platforms. As discussed in \cite{jukan2017evolution} AI-based techniques can also be used to \emph{learn} and automate the process orchestration workflow for incoming intent requests, while retaining the possibility of human intervention in the control loop. 
	
	The widespread deployment of SDN, coupled with a telemetry-driven feedback loop, would reduce the barrier for automating network control operations. AI-based techniques presented in Section \ref{subsec:connection} can then be used to optimize RWA/RMLSA/RMCSA problems, performing in-operation planning to optimize resources based on existing optical service demands. As AI-based techniques in the area of network control and management mature, we can also expect advanced applications such as preemptive relocation of services away from a predicted failure in the network, a topic in which first steps have already been taken \cite{siracusa2014proactive, fernandez2014demonstration, vela2017ber, vela2017combining}. As outlined in Section \ref{sec:role-AI-networks}, network virtualization and reconfiguration could also take advantage of AI-based techniques by utilizing traffic prediction and classification, with or without SDN controller in the loop. Nevertheless, the introduction of flex-grid devices, such as bandwidth-variable sliceable transponders and WSSs, and space division multiplexing techniques based on multi-core or few-/multi-mode fibers have further increased the computation complexity of AI-enabled RWA/RMLSA/RMCSA optimization solutions \cite{jukan2017evolution}. Nevertheless, AI-based techniques, especially deep learning, will play a major role in optical network planning and reconfiguration, since optical devices are expected to imbibe more autonomous programmability features in the next decade.

	\subsection{Efficient Joint Operation of Networks and Computing Resources}
	Emerging paradigms like the Internet of things (IoT), Industry 4.0 or the tactile Internet \cite{maier2016tactile}, impose stringent requirements on networks, such as low latency, and high bandwidth, availability and security, thus posing a significant challenge. The combination of 5G mobile communications systems with high-speed fault-tolerant fiber backhaul infrastructures will be key enabling technologies for these networks \cite{maier2016tactile}. End-to-end latency for some applications can be limited to few milliseconds (e.g., 1 ms for tactile internet). Thus, the distance between the edge and computing resources must be limited to some tens of kilometers \cite{wong2017predictive}, and a decentralized service platform architecture based on Mobile Edge Computing (MEC) or Fog Computing (FC) is required.  However, the integration of  various computing paradigms (MEC, Fog  and cloud) involves the development of integrated resource management, task allocation and failure handling techniques, to name just a few. Therefore, the joint allocation of computing and networking resources (also including inter-datacenter networking) is receiving increasing attention.
	
	AI is expected to play a key role to facilitate efficient joint operation of network and computing devices, performing tasks like virtual network function (VNF) distribution, task allocation, predictive caching and interpolation/extrapolation of human actions, and thus enhancing performance and providing better support for IoT and tactile Internet applications. For instance, along this line, a recent work by Wong \textit{et al}.\ \cite{wong2017predictive} has proposed a novel tactile Internet capable PON and a dynamic wavelength and bandwidth allocation method, which incorporates a mechanism to predict the traffic load to vary the number of active wavelength channels in the network, and prioritize the transmission of tactile Internet traffic (vs. other traffic) to comply with delay requirements. Due to the huge expansion of IoT applications and services, we envision more advances to come along this line in the next years.
	
	\subsection{Applications in On-Chip Networks}
	Finally, another promising field of application in our opinion is that of on-chip networks. Today, major tech companies like Google, Microsoft, Intel, etc. are pushing chip manufacturing to enable AI computing on a single chip. On-chip networks provide communication substrates for the constantly increasing number of cores on a single chip. Generally, on-chip networks are designed to enable efficient computing in less time while consuming less power. Though it is early days in chip manufacturing to bring AI-enabled on-chip networks in market, recently Google has come up with its own AI chip, known as Tensor Processing Unit (TPU), which allows TensorFlow (a deep neural network software) to run. Similarly, Intel announced  an experimental chip called ``Loihi", which is designed based on  neuromorphic technology that uses neurons instead of logic gates. AI algorithms are known for time and computation complexity, thus requiring multiple central and graphical processing units, therefore AI-enabled optical network on-chip (ONoC) is an alternative to the electronic NoC which will further reduce the power consumption and computation time. For instance, Gu \textit{et al}.\ \cite{Gu2017Time} have recently proposed an optical network on-chip architecture by combining TDM and WDM technologies. This novel architecture allows to solve the blocking problem faced by previously proposed optical circuit-switching (OCS) based ones \cite{Hendry2010Silicon}. Specifically, as an aspect of special interest for the subject of this article, the number of wavelength groups and time slots is optimized by using a genetic algorithm, which helps TDM-WDM-based ONoC to outperform results from equivalent proposals based on OCS-mesh.
	
	\section{Summary} \label{sec:conclusion}
	
	This paper has provided a comprehensive survey of the current research within the application of Artificial Intelligence (AI) techniques in optical networks, as well as an overview of opportunities and challenges arising in this context. 
	
	In order to provide the reader with a clear and general vision of the numerous techniques and methods that make up this scientific discipline, we have first described those AI subfields that have been successfully employed in optical networking: (1) search methods and optimization theory, (2) game theory, (3) knowledge-based, reasoning and planning methods, (4) statistical models, (5) decision-making algorithms, and (6) learning methods, and have classified relevant literature in the diagram shown in Figure \ref{figure:AI-opticalnetworks}.
	
	Later on, we have extensively reviewed the application of these techniques in order to improve the efficiency of both optical transmission (Table \ref{table:AI-physical}), and the design and control of optical networks (Tables \ref{table:AI-networks-I} and \ref{table:AI-networks-II}). Specifically, in relation to optical transmission, we have addressed the suitability of AI techniques when dealing with the characterization and operation of transmitters, EDFAs and receivers, as well as for performance monitoring, mitigation of nonlinearities, and for QoT estimation, which is particularly relevant in impairment-aware optical network operation. In many cases, AI techniques have proven to be more efficient than classical approaches, which are typically limited by the complexity, lack of adaptability, and/or scalability of the deterministic or semi-analytical models on which they rely. AI techniques provide similar advantages in issues related to the control and design of optical networks. The thorough review of the state of the art in this matter has included numerous applications in the following fields: optical network planning, connection establishment (i.e., the pursuit of the optimal solution for the RWA, RSA, RMLSA or RMCSA problems), network reconfiguration, software-defined networking and applications in specific types of networks such as OBS networks, Passive Optical Networks and data centers.
	
	In general, AI techniques' ability to find optimal or near-optimal solutions in highly complex scenarios (i.e., problems with very high dimensionality, fed with huge amounts of data) without the need to develop exhaustive analytical or semi-analytical models, makes them essential to meet the challenge posed by the increased complexity and dynamism of current optical communication networks, which need to face, among other impairments, the physical limitations imposed by noise and nonlinear distortions. No less important is AI techniques' ability to adapt to changing conditions, to learn from them and to propose solutions to unexpected situations or cases, turning them into promising candidates to, for example, reconfigure these dynamic networks in short time scales in order to meet changing demand patterns.
	
	Last but not least, we have described a number of opportunities and challenges that we envision in the area of optical networking. In the context of optical transmission, we expect AI will continue playing a crucial role in supporting technologies like space division multiplexing, multimode/multicore fibers, advanced modulation formats and constellation shaping, etc. Notwithstanding, the role that AI can play in attack and intrusion detection in optical networks seems of special relevance and has not yet been significantly explored to the best of our knowledge. Furthermore, the automation of network management operations, especially in the current context of networks becoming increasingly heterogeneous, and the efficient joint operation of networks and MEC/fog/cloud computing resources, together with on-chip networking are, in our opinion, fields where AI can be decisive.

	\section*{Acknowledgment}
	This work has been partially supported by the Spanish Ministry of Economy and Competitiveness (TEC2014-53071-C3-2-P, TEC2015-71932-REDT).

	\section*{References}
	
	\bibliography{AIbib}
	
\end{document}